\documentclass{article}

\usepackage{arxiv}

\usepackage[utf8]{inputenc} 
\usepackage[T1]{fontenc}    
\usepackage{hyperref}       
\usepackage{url}            
\usepackage{booktabs}       
\usepackage{amsfonts}       
\usepackage{nicefrac}       
\usepackage{microtype}      
\usepackage{lipsum}
\usepackage{amsmath}
\usepackage{multirow}
\usepackage{cite}
\usepackage{graphicx}
\graphicspath{ {./images/} }
\usepackage{graphicx}
\usepackage{subfigure}
\usepackage[colorinlistoftodos]{todonotes}
\usepackage{comment}

\title{Lesion2Vec: Deep Metric Learning for Few Shots Multiple Abnormality Recognition in Wireless Capsule Endoscopy Video}

 \author{
  Sodiq ~Adewole \\
  Systems \& Information Engineering \\
  University of Virginia, Charlottesville, VA 22903 \\
  
  \And
  Philip ~Fernandez \\
  Department of Pediatrics, School of Medicine \\ 
  University of Virginia, Charlottesville, VA, USA \\
  
  \And
  James ~Jablonski \\
  Systems \& Information Engineering \\
  University of Virginia, Charlottesville, VA 22903 \\
    
  \And
  Michelle Yeghyayan \\
  Department of Pediatrics, School of Medicine \\ 
  University of Virginia, Charlottesville, VA, USA \\
   
    \And
  Michael ~Porter \\
  Systems and Information Engineering \\
  University of Virginia, Charlottesville, VA 22903 \\
  
  \And
  Andrew ~Copland \\
  Department of Pediatrics, School of Medicine \\ 
  University of Virginia, Charlottesville, VA, USA \\

    \And
  Sana ~Syed \\
  Department of Pediatrics, School of Medicine \\ 
  University of Virginia, Charlottesville, VA, USA \\
  
  \And
  Donald ~Brown \\
  Data Science Institute \\
  University of Virginia, Charlottesville, VA 22903 \\
}

\begin{document}
\maketitle

\begin{abstract}
  Effective and rapid detection of lesions in the Gastrointestinal tract is critical to how promptly gastroenterologist can respond to some life-threatening diseases. Wireless Capsule Endoscopy (WCE) has revolutionized traditional endoscopy procedure by allowing gastroenterologists visualize the entire GI tract non-invasively. Once the tiny capsule is swallowed, it sequentially capture images of the GI tract at about 2 to 6 frames per second (fps). A single video can last up to 8 hours producing between 30,000 to 100,000 images. Automating the detection of frames containing specific lesion in WCE video would relieve gastroenterologists the arduous task of reviewing the entire video before making diagnosis. While the WCE produces large volume of images, only about 5\% of the frames contain lesions that aid the diagnosis process. Convolutional Neural Network (CNN) based models have been very successful in various image classification tasks. However, they suffer excessive parameters, are sample inefficient and rely on very large amount of training data. Deploying a CNN classifier for lesion detection task will require time-to-time fine-tuning to generalize to any unforeseen category. In this paper, we propose a metric-based learning framework followed by a few-shot lesion recognition in WCE data. Metric-based learning is a meta-learning framework designed to establish similarity or dissimilarity between concepts while few-shot learning (FSL) aims to identify new concepts from only a small number of examples. We train a feature extractor to learn a representation for different small bowel lesions using metric-based learning. At the testing stage, the category of an unseen sample is predicted from only a few support examples, thereby allowing the model to generalize to a new category that has never been seen before. We demonstrated the efficacy of this method on real patient capsule endoscopy data with additional category during testing that was not part of our training. Experiments were conducted on the impact of the number of support samples and different CNN architecture models on the recognition performance. The experiment results show that this approach is effective in few-shot lesion recognition in WCE data.
\end{abstract}


\section{Introduction}
Wireless Capsule Endoscopy allows non-invasive visualization of the entire gastrointestinal tract including the small-bowel region by the gastroenterologist for various disease diagnosis. Traditional upper and lower endoscopy procedure have limited visibility as they can only allow visualization of the upper and lower GI tract leaving the small bowel region inaccessible. A tiny capsule swallowed by the patient captures images at about 2 - 6 \footnote{https://www.medtronic.com/covidien/en-us/products/capsule-endoscopy/pillcam-sb-3-system.html} frames per second (fps) as it is propelled down the GI tract through intestinal peristalsis. A single WCE examination could last up to 8 hours producing between 30,000 to 100,000 images compiled as a video. The collected images are subsequently transferred to a work station where they are reviewed and analysed frame-by-frame by an expert gastroenterologist for diagnosis. Usually, only about 5\% of the entire video contain visual features that aid the diagnosis process. While WCE is an innovative technology, automating detection of lesions is critical to increase its clinical application.

For over two decades, automatic detection of lesions in WCE data has received much attention from researchers and several approaches have been proposed in literature \cite{rahim2020survey, miaou2009multi, kodogiannis2008neuro, sainju2014automated, van2009capsule, mamonov2014automated}. 

Recently, Convolutional Neural Networks (CNN) based model have gained significant attention and are currently the state-of-the-art models for image classification and object detection tasks \cite{simonyan2014very, he2016deep, krizhevsky2012imagenet, russakovsky2015imagenet}. While CNN based models have demonstrated great success, they generally require a large amount of labeled examples for training and are sample inefficient \cite{zhang2021few}. Generally, due to the required expertise, obtaining labels for medical data can be very difficult. Moreover, while a single endoscopy examination may contain multiple lesions, only about 5\% of the entire CE video offer informative content that aid gastroenterologist in their diagnosis, producing far more examples of normal frames than the useful abnormal ones.

Another unique property of WCE data is the approach to data collection. Gastroenterologist are only able to obtain sample data from patients who visit for a specific problem. Traditional classification model trained based on currently available data will require fine-tuning when new cases of unseen categories arise. Thereby calling for another round of sample collection in such large quantity as would ensure the generalizability of the model to the new category. Such time-to-time fine-tuning for every new lesion category would ultimately limit the adoption of the system in real life clinical situation.

In contrast, humans possess a remarkable ability to learn a new concept from only a few instances and quickly generalize to new circumstances \cite{zhang2021few}. Given one or few template examples, humans are able to generalize to new circumstances. We employed similar concept to automatically recognize new category of lesion in WCE data.


\subsection{Related Work}
\label{sec:related_work}
Prior efforts to automate the analysis and detection of lesions in capsule endoscopy data can be broadly grouped into;
\begin{itemize}
\item Abnormal frame or outlier detection framework \cite{gao2020deep, miaou2009multi, zhao2010abnormality}. Broad categorization into normal/abnormal helps reduce redundancy rate caused by large number of normal frames and also minimize the review time and effort needed by the gastroenterologist to make diagnosis. However, this approach does not offer any granular information that specifically help identify characteristic lesion in the images. They also still require the gastroenterologist to review the frames before making decision.
\item Informative/key frame extraction \cite{chen2016wireless, emam2015adaptive, iakovidis2010reduction, tsevas2008automatic}: This approach is otherwise known as the video summarization framework where the model learns to extract key frames (lesion-containing or abnormal) from entire video sequence.
\item Models to detect specific abnormality or lesion such as bleeding in \cite{sainju2014automated}, polyp in \cite{van2009capsule, mamonov2014automated, hwang2010polyp, yuan2015improved}, ulcer \cite{yuan2015saliency}, and angioectasia \cite{tsuboi2020artificial, pogorelov2018deep, deeba2018saliency}.
\end{itemize}

We will review each of these prior works in more detail in the following.

In \cite{miaou2009multi} Miaou et al. propose a four-stage classification model based on low-level Hue-Saturation-Intensity (HSI) features followed by fuzzy-c means clustering analysis to separate images carrying different abnormalities in such step-wise manner. The final stage is a neural network model that discrimininate normal from abnormal frames. In \cite{mewes2012semantic} Mewes et al. applied similar multi-stage technique to extract quality frames by removing over-/under-expose images as well as images with significant non-tissue areas.
Using color histogram of images, \cite{kodogiannis2008neuro} employed fuzzy neural model which combines fuzzy systems and artificial neural networks to detect abnormal lesions in CE images.

Zhao et al. \cite{zhao2010abnormality} proposed a temporal segmentation approach based on adaptive non-parametric key-point detection model using multi-feature extraction and fusion. The aim of their work was not only to detect key abnormal frames using pairwise distance, but also to augment gastroenterologist's performance by minimizing the miss-rate and thus, improving detection accuracy. \cite{sainju2014automated} proposed to detect bleeding regions in frames by computing statistical features of the first order histogram probability of the three color channels (RGB) in the images before passing the computed features to a neural network to discriminate bleeding from non-bleeding frames.`In order to develop scalable models, such low level hand-crafted feature extraction method may not scale properly.

\cite{mamonov2014automated} Mamonov et al. propose a model for colorectal polyp detection based on a binary classification using geometric analysis and texture content of the frames. Their model achieve 47\% sensitivity and 90\% specificity. Similarly, Hwang et al. \cite{hwang2010polyp} propose a polyp detection model by first segmenting the affected region using Gabor texture features and the applying K-means clustering. The resulting geometric information is then used to identify frames containing polyp. Yixuan et al. \cite{yuan2015improved} proposed a bag of feature (BoF) technique using integration of multiple features such as texture features, scale-invariant feature transform (SIFT), complete local binary pattern (CLBP) with visual words to automatically detect polyp in WCE image. While SIFT remains the baseline feature for traditional image analysis, CNN based models achieve significantly improved performance in complex geometric and lighting conditions which is typical of CE data.

Traditional low-level feature extraction techniques have been well explored in CE image analysis. However, little attention has been given to CNN-based models. For example, Akiyoshi et al. \cite{tsuboi2020artificial} work uses CNN-based model - Single Shot Multibox Detector - to automatically detect frames with angioectasia in CE images while \cite{deeba2018saliency} proposed a saliency-based unsupervised method for the same task. \cite{pogorelov2018deep} combined deep learning and handcrafted features for polyp detection. \cite{gao2020deep} applied CNN-based model in a semi-supervised context to detect frames with abnormality. The main drawback of CNN-based models is that they typically require vast quantity of labelled data and suffer from poor sample efficiency, which excludes many application where data is typically rare or expensive \cite{liu2018deep}. Given the volume of frames generated in WCE data, the cost of obtaining expert label for every frame across multiple patient is generally prohibitive. 

Given large amount of labelled data, state-of-the-art performance can be achieved by a CNN based classifier model on different lesion categories. However, given the peculiarity of WCE data, achieving 100\% accuracy on limited category is not sufficient for real world clinical application. This can be further exacerbated by the far more normal example frames observed in each video. Making obtaining diverse and sufficient examples of every new lesion category \cite{zhang2021few} difficult.
We propose to tackle this problem using the Metric-based learning and few-shots classification. To the best of our knowledge, we are not aware of any prior efforts that has explored few-shot learning on capsule endoscopy image analysis.

Our contributions are highlighted as follows:
\begin{itemize}
    \item This is the first work to propose state-of-the-art few-shot learning in capsule endoscopy image analysis field. Our experimental evaluation shows that it is possible to learn much information for a new category of lesion from just a few examples.
    \item We conduct extensive experiment to investigate factors that affect performance including which CNN architecture produces better performance and the impact of support samples (shots) on the different lesion categories.
\end{itemize}

\subsection{Outline}
The remaining part of the paper is organized as follows. Section \ref{sec:background} reviews the principles and basic framework of Metric and few-shot learning. Section \ref{sec:method} presents the application of few-shot learning for lesion recognition in WCE data. Section \ref{sec:experiments} mainly covers the dataset and experimental results. Section \ref{sec:conclusion} summarizes the paper and key direction for future works.
\section{Theoretical Background}\label{sec:background}
\subsection{Overview of Few-shot learning}
Few-shot learning is a special case of meta-learning where we aim to learn new concepts from a limited number of labeled samples and quickly adapt to unforeseen tasks. The few-shot learning task is an extension of the single-shot learning \cite{koch2015siamese} framework where the discriminating potential of the learned embedding space is evaluated. For single-shot learning, given a test image $x_{i}$, that we wish to classify into $1$ of $C$ classes, we are also given one example image $\{x^{c}\}^{C}_{1}$ from each category. We can then query the network to compute the embedding for $\{x_{i}$, $x_{c}\}$ $\forall$ $c = 1, ..., C$. 
We then predict the class corresponding to the minimum distance from the test image. 

\begin{equation}
    C^{*} = \operatorname*{argmin}_c d \{f(x_{i}), f(x_{c})\}
\end{equation}
where $d$ is the distance metric between $f(x_{i})$ and each of $f(x^{c})$.

In few-shot learning, there are more than one example of each category. In this case, it is possible to learn a distribution over the embedding space for each category.

\subsection{Overview of Metric-based Learning}
Metric-based learning is one of the family of approaches used in few-shot learning \cite{vanschoren2018meta} where the model learn a representation of the current task such that, given a few support instances, it is able to generalized to an unseen task \cite{zhang2021few}. Metric based learning is designed to maximize the inter-class distance between embedding features belonging to different class while simultaneously minimizing the distance between embedding features belonging to the same class. Some of the proposed architecture in metric-based learning are the Siamese Neural Network \cite{bromley1994signature} and the Triplet Network \cite{hoffer2015deep}.

\begin{figure}[h]
  \centering
  \includegraphics[trim={1.1cm 17.1cm 9.1cm 5.1cm},clip,width=0.5\linewidth]{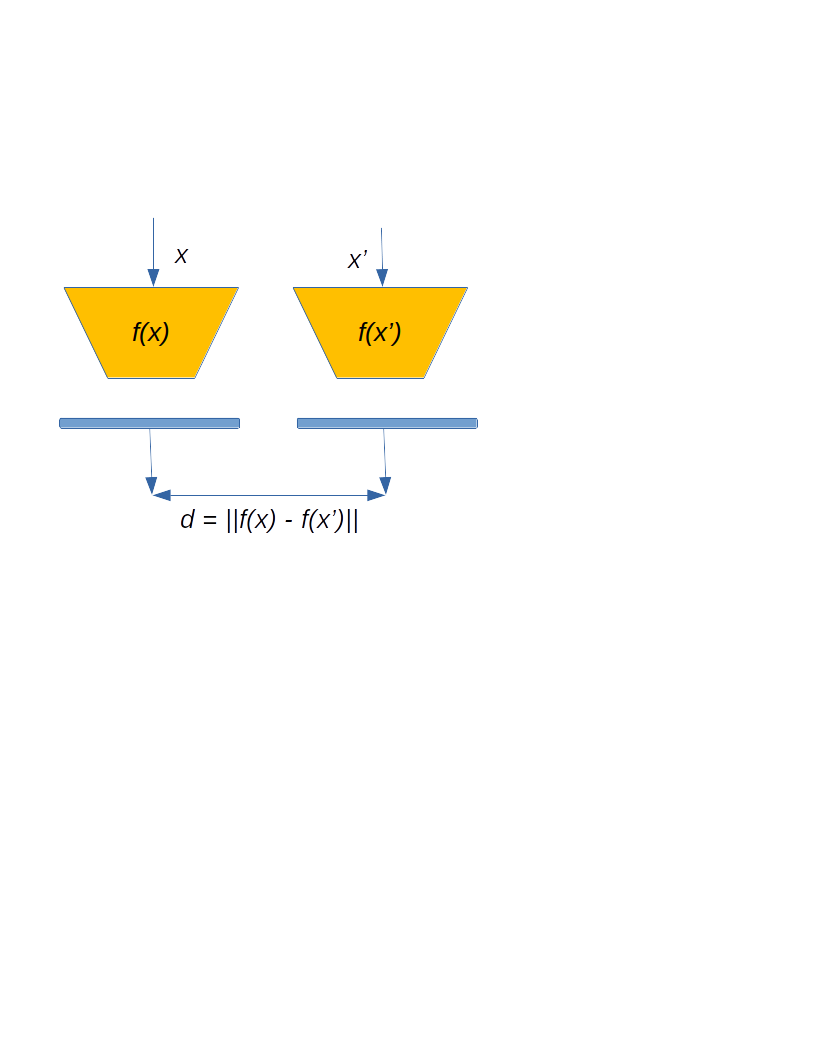}
  \caption{Siamese Network}
  \label{siamese_network}
\end{figure}

\subsubsection{Triplet Network}
The Triplet Network (TN) proposed by \cite{hoffer2015deep} was inspired by Siamese Neural Network (SNN), proposed in \cite{bromley1994signature}. The SNN was proposed to first solve signature verification problem as an image matching task. As shown in Fig. \ref{siamese_network}, the SNN consist of two identical sub-networks that are trained simultaneously. The networks share the same set of parameters and are joined at their outputs by a joining neuron. The networks extract features from a pair of input images of different categories while the joining neuron measures the distance between the two feature vectors. Learning in the twin networks is done with distance-based metric. \cite{bromley1994signature} proposed the contrastive loss which computes the euclidean distance between every pair of inputs $d = {x_{1}, x_{2}}$. During training, the SNN is optimized to minimize the distance between pairs of vector representing inputs from the same class and increase the distance between vector representation of inputs from different classes. Subsequently, \cite{hoffer2015deep} proposed the triplet loss function that combines two contrastive losses between an anchor and an input from same class (positive) and the anchor with an input from a different class (negative) to form a triplet. This architecture is shown in figure \ref{triplet}. The TN and SNN have been successful in other domains on face verification and recognition problems \cite{Schroff_2015_CVPR, Taigman_2014_CVPR} where the models directly learns a mapping from face images to compact Euclidean space. \cite{koch2015siamese} also reported effectiveness of the network on character recognition problem. 
\begin{equation} \label{triplet}
    \|x_{i}^{a} - x_{i}^{p}\|_{2}^{2} + \alpha < \|x_{i}^{a} - x_{i}^{n}\|_{2}^{2},
    \hspace{3pt} \forall \hspace{3pt} (x_{i}^{a}, x_{i}^{p}, x_{i}^{n}) \hspace{5pt} \in \hspace{3pt} \mathcal{T}
\end{equation}

where $\alpha$ is a margin that is enforced between positive and negative pairs to prevent the network from learning a trivia solution. $\mathcal{T}$ is the set of all possible triplets in the training set and has cardinality $N$. 
In this paper we applied the triplet loss function in training the network parameters.

\subsubsection{Triplet loss}
\begin{figure}[h]
  \centering
  \includegraphics[trim={2.1cm 10.0cm 2.1cm 1.1cm},clip,width=0.6\linewidth]{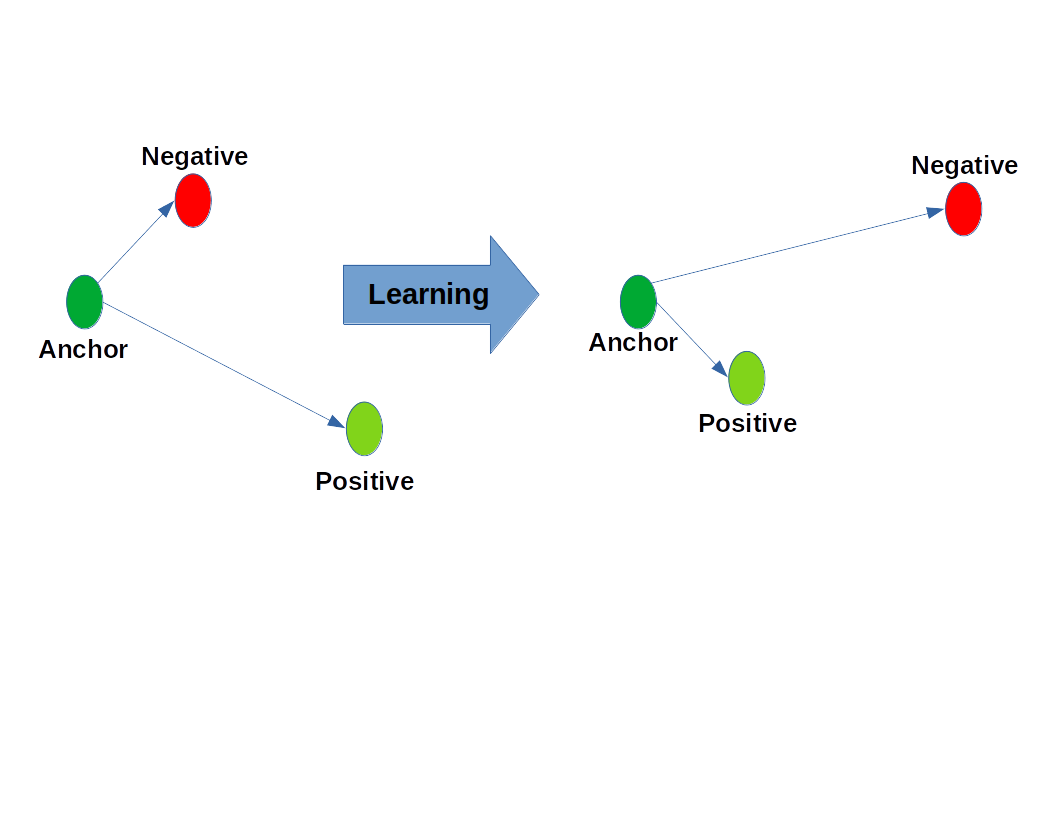}
  \caption{Triplet Loss}
  \label{triplet_loss}
\end{figure}

The triplet loss function shown in eq. (\ref{eq:triplet}) optimizes the parameters of the network to minimize the distance between an anchor and a positive instance through projection to a single point in the embedding space. It simultaneously maximizes the distance between the anchor and a negative instance as shown in figure \ref{triplet_loss}. The embedding function, represented as $f(x) \in \mathcal{R}^D$ parameterized by $\theta$ extracts features from the input image $x$ to produce a $D$-dimensional feature vector. The loss ensures that a frame $x_{i}^{a}$ (\textit{anchor}) with a specific lesion is closer to all other frames $x_{i}^{p}$ with the same lesion than to other frames $x_{i}^{n}$ (\textit{negative}) with a different lesion or normal tissue.

\begin{equation}\label{eq:triplet}
    \sum_{i=1}^{N}\left[\|f(x_{i}^{a}) - f(x_{i}^{p})\|_{2}^{2} - \|f(x_{i}^{a}) - f(x_{i}^{n})\|_{2}^{2} + \alpha \right]_{+}
\end{equation}
where $f$ represents the function that the CNN model learns.

\section{Proposed Method}
\label{sec:method}
WCE videos have variety of challenging characteristics. Due the complex structure of the GI tract frames from the capsule camera, the images suffer from uneven illumination, low resolution, variable focal sharpness, and high compression ratio. Some of the video frames contain highly light reflections or maybe out-of-focused because of peristalsis as well as completely accidental WCE movements through the peculiar GI tract. Moreover, non-lesion frames can show deceptive structures such as bubbles, extraneous, food items, fecal matter, turbid fluids, gastric/intestinal juices, etc.

\begin{figure}[h]
    \centering
    \begin{subfigure}
        \centering
        \includegraphics[trim={1.1cm 1.1cm 1.1cm 1.1cm},clip,width=0.15\textwidth]{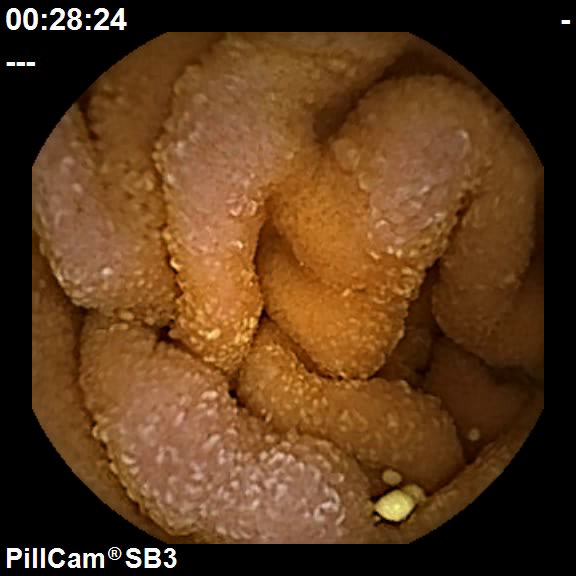}
        \includegraphics[trim={1.1cm 1.1cm 1.1cm 1.1cm},clip,width=0.15\textwidth]{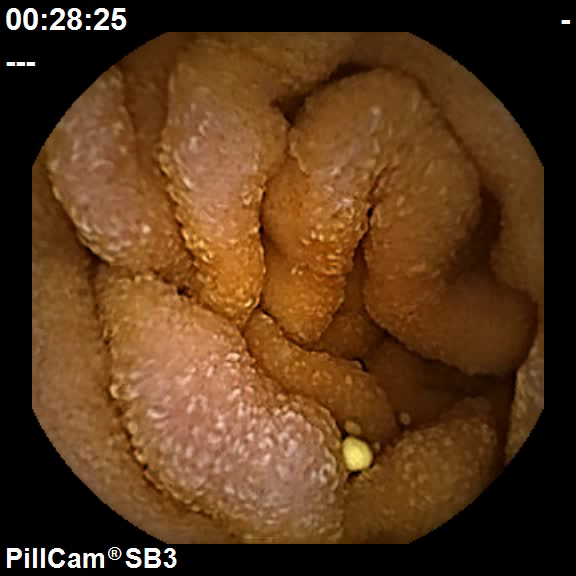}
        \includegraphics[trim={1.1cm 1.1cm 1.1cm 1.1cm},clip,width=0.15\textwidth]{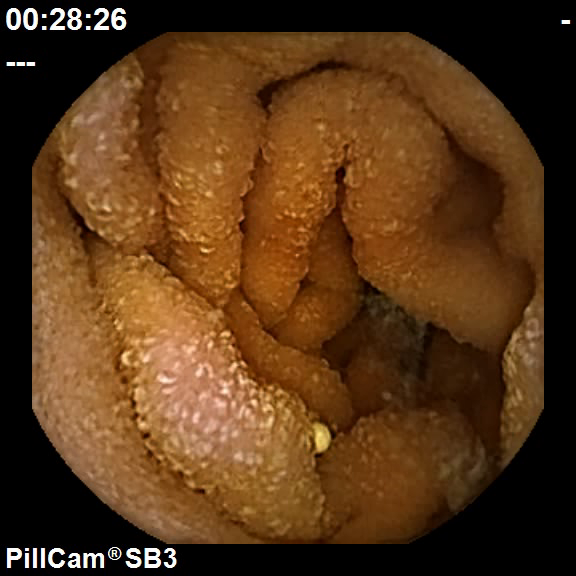}
        \includegraphics[trim={1.1cm 1.1cm 1.1cm 1.1cm},clip,width=0.15\textwidth]{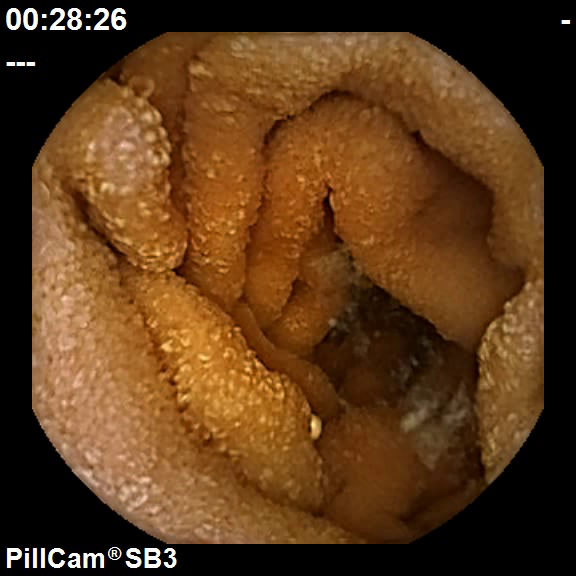}
    \end{subfigure} \\
    \begin{subfigure}
        \centering
        \includegraphics[trim={1.1cm 1.1cm 1.1cm 1.1cm},clip,width=0.15\textwidth]{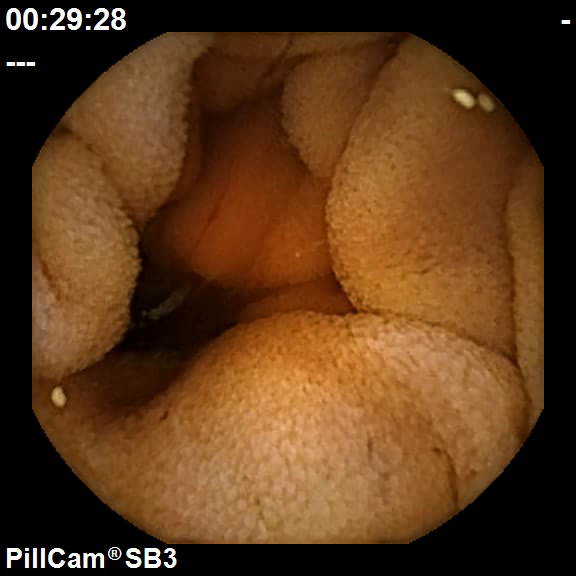}
        \includegraphics[trim={1.1cm 1.1cm 1.1cm 1.1cm},clip,width=0.15\textwidth]{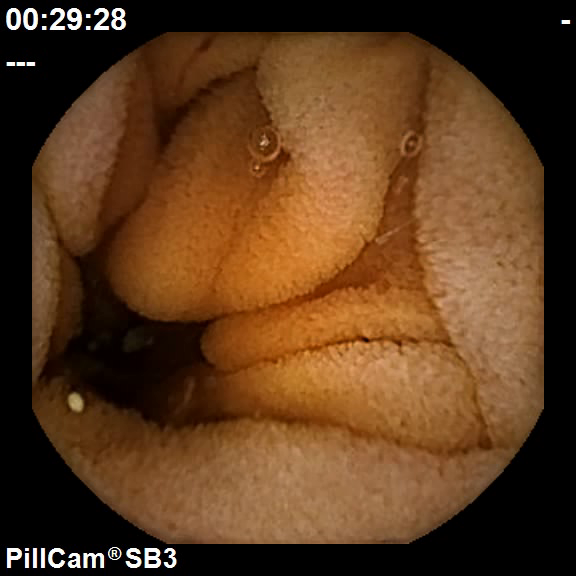}
        \includegraphics[trim={1.1cm 1.1cm 1.1cm 1.1cm},clip,width=0.15\textwidth]{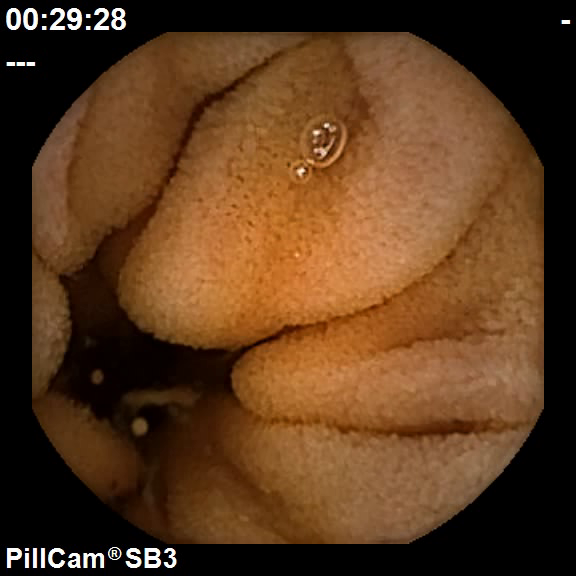}
        \includegraphics[trim={1.1cm 1.1cm 1.1cm 1.1cm},clip,width=0.15\textwidth]{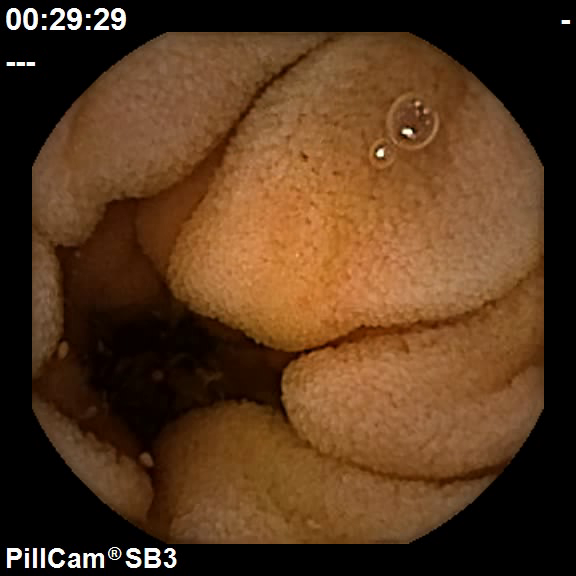}
    \end{subfigure} \\
    \begin{subfigure}
        \centering
        \includegraphics[trim={1.1cm 1.1cm 1.1cm 1.1cm},clip,width=0.15\textwidth]{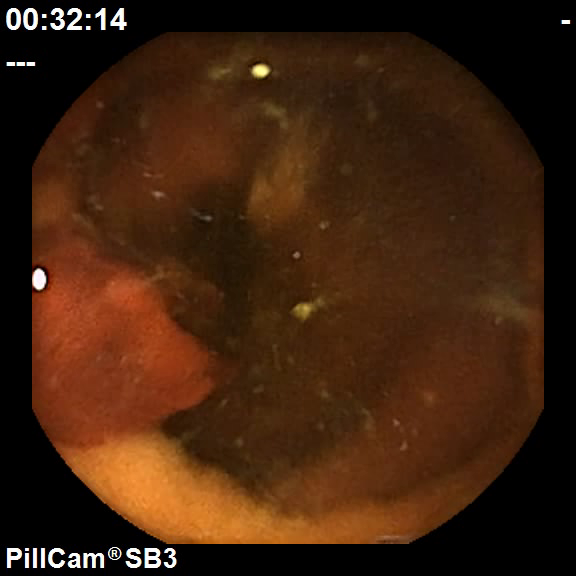}
        \includegraphics[trim={1.1cm 1.1cm 1.1cm 1.1cm},clip,width=0.15\textwidth]{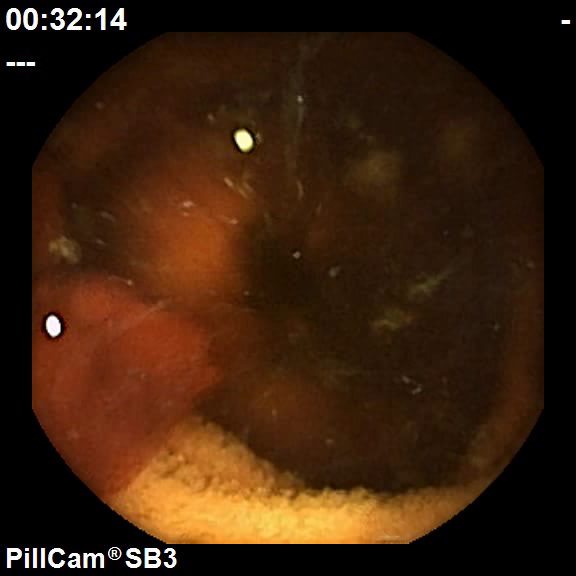}
        \includegraphics[trim={1.1cm 1.1cm 1.1cm 1.1cm},clip,width=0.15\textwidth]{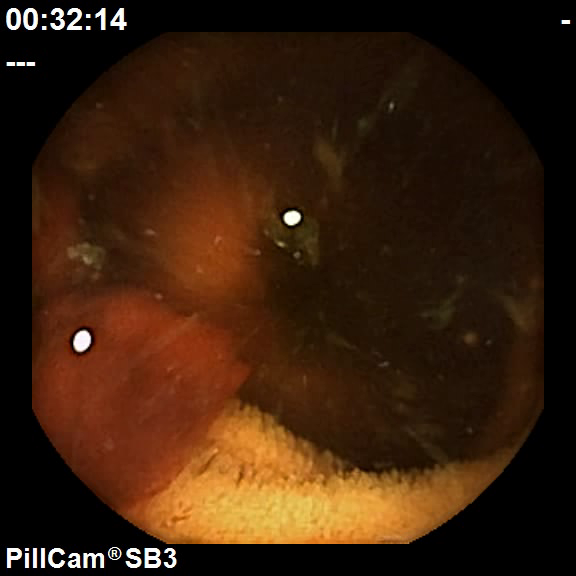}
        \includegraphics[trim={1.1cm 1.1cm 1.1cm 1.1cm},clip,width=0.15\textwidth]{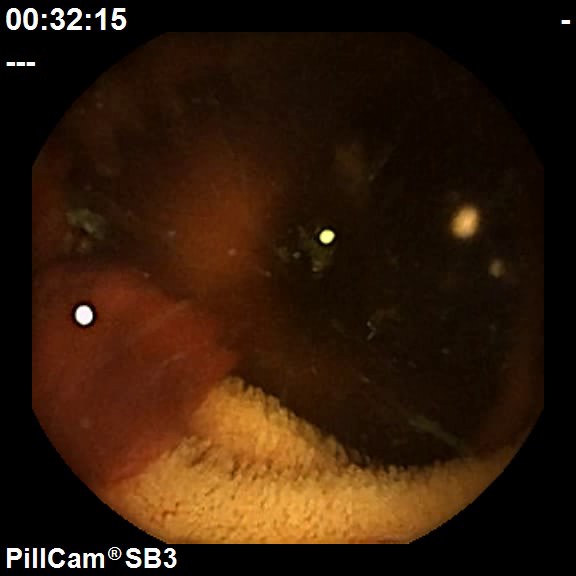}
    \end{subfigure} \\
    \begin{subfigure}
        \centering
        \includegraphics[trim={1.1cm 1.1cm 1.1cm 1.1cm},clip,width=0.15\textwidth]{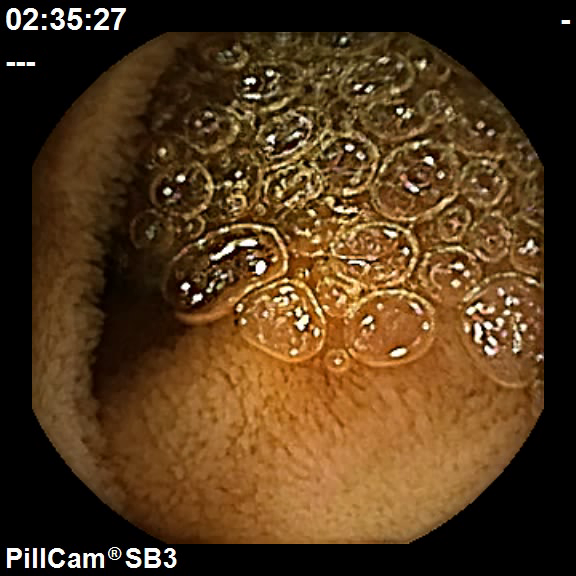}
        \includegraphics[trim={1.1cm 1.1cm 1.1cm 1.1cm},clip,width=0.15\textwidth]{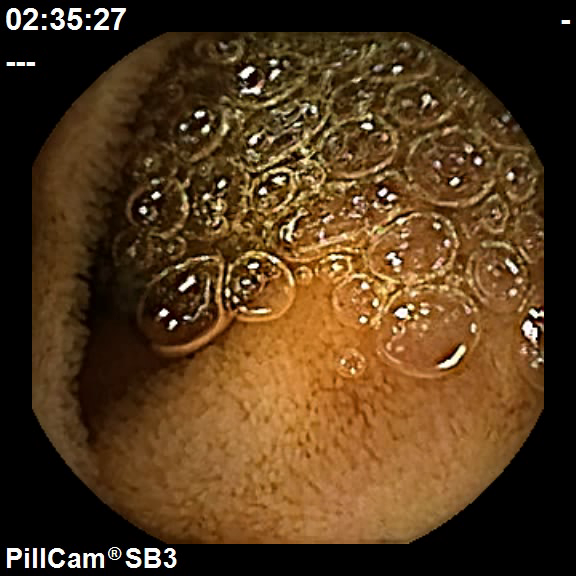}
        \includegraphics[trim={1.1cm 1.1cm 1.1cm 1.1cm},clip,width=0.15\textwidth]{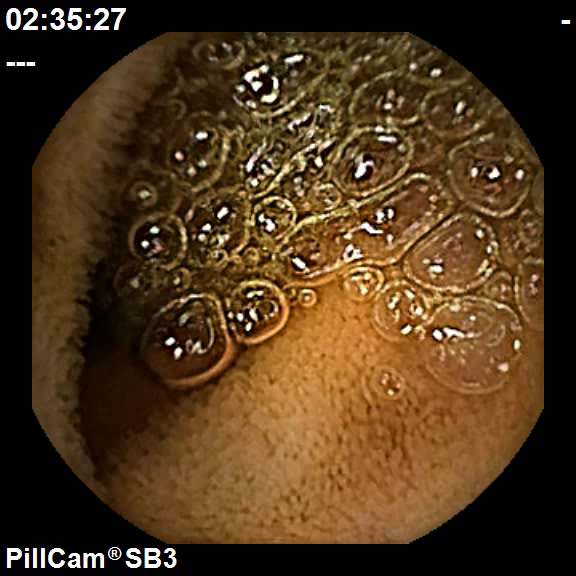}
        \includegraphics[trim={1.1cm 1.1cm 1.1cm 1.1cm},clip,width=0.15\textwidth]{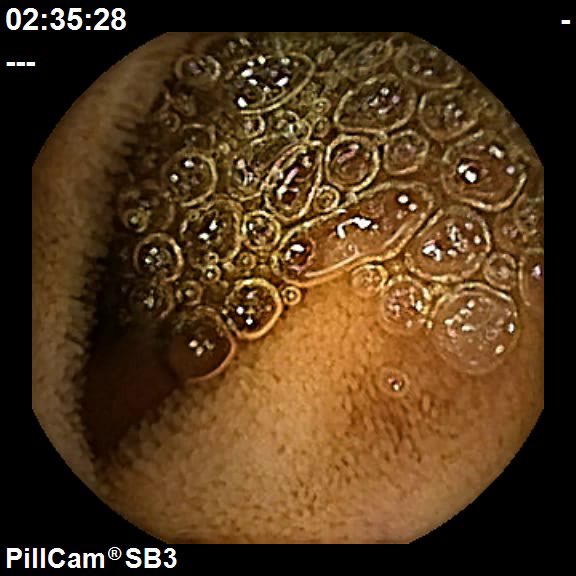}
    \end{subfigure}
\caption{Samples from WCE frames the small bowel showing whipples; Normal; Diffuse bleeding; and Bubbled frames}
\label{sample_frames}
\end{figure}

\begin{figure}[h]
  \centering
  \includegraphics[width=0.8\linewidth]{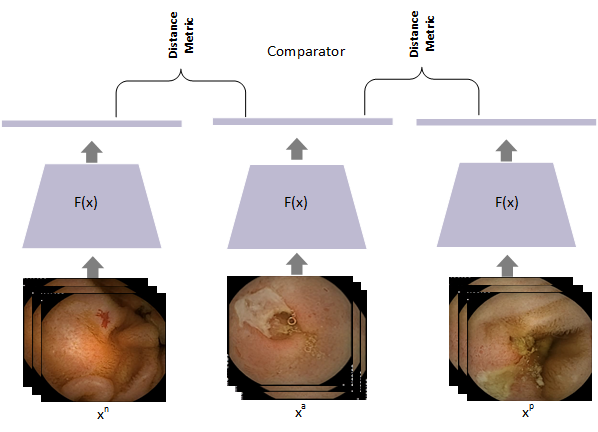}
  \caption{Triplet Network}
  \label{fig:triplet_network}
\end{figure}

Here we propose deep metric learning to perform few-shot classification on WCE images. The framework for the proposed architecture is shown in Figure \ref{fig:triplet_network}

\begin{figure}[h]
    \centering
    \includegraphics[width=0.8\linewidth]{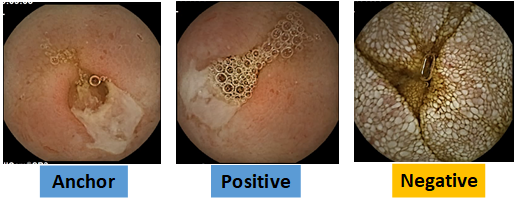}
    \caption{Example triplet with an anchor, positive and a negative}
    \label{fig:triplet}
\end{figure}

Specific example selection of an anchor, positive and negative instances in shown in Figure \ref{fig:triplet}

\textbf{Step 1} (Pre-processing): After exporting the raw video files from the RapidReader software, we processed all videos into frames. Since the gastroenterologist is usually more interest in the small bowel region, we only focus on images of the small bowel. Each frame was pre-processed to trim off the uninformative boundary region. 

\textbf{Step 2} (Sampling): Each batch sampling involve a triplet of anchor, positive and negative instances \cite{Schroff_2015_CVPR}. In WCE, the anchor is randomly initialized to one of the training categories containing certain lesion (e.g. ulcer as in fig. \ref{fig:triplet}), a positive instance is another frame with the same lesion while a negative example is a frame with a different lesion or from a normal category. During training, we used the triplet loss (implemented in PyTorch\cite{NEURIPS2019_9015}) to optimize the model parameters to force the embedding of similar frames into the same region of the feature space, such that the squared distance between the anchor and a positive is minimal while simultaneously maximizing the distance between the anchor and any negative instance. 

\textbf{Step 3} (Forward Pass): We passed the triplet sample to the network to compute the embedding for each frame. We applied the euclidean distance to compute the distance between the triplet. For each epoch, we compute the triple loss based on Eq. \ref{triplet_loss} as implemented in the PyTorch framework. 

\textbf{Step 4} (Back-propagation): Parameters of the network were updated using back-propagation to gradually reduce the loss. We experimented with different optimizers but found the Stochastic Gradient Descent to work best. Our learning rate was set to 0.001 and each model was trained for 150 epochs. Each frame was embedded to a fixed feature vector of dimension 128.

\subsection{Evaluation Criteria}
We evaluated the performance of the model on few-shot classification task based on different criteria: Precision, Recall, F1-score and Accuracy. The evaluation metrics are computed based on the number of correctly identified lesion (true positive, TP); the number of correctly identified frames as containing a different lesion for each class(true negative, TN), the number of missed frames containing a particular lesion (false negative, FN); and the number of normal frames wrongly identified as containing a particular lesion (false positives, FP). We compute the evaluation metrics based on the equations below:

\begin{equation}
    Precision = \frac{TP}{TP + FP}\\
\end{equation}

\begin{equation}
    Recall = \frac{TP}{TP + FN}\\
\end{equation}

\begin{equation}
    F1 = 2 \left(\frac{Precision x Recall}{Precision + Recall}\right)\\
\end{equation}

\begin{equation}
    Accuracy = \frac{TP + TN}{TP + TN + FP + FN}
\end{equation}
\section{Experiments}
\label{sec:experiments}
This section details the makeup of our dataset and experimental steps.
\subsection{Dataset Introduction}
Our dataset consist of real patients capsule endoscopy data collected under supervision of expert gastroenterologist. The dataset were collected from 52 patients using SB3 PilCam  \footnote{https://www.medtronic.com/covidien/en-us/products/capsule-endoscopy/pillcam-sb-3-system.html}. Using the RapidReader software \footnote{https://www.medtronic.com/covidien/en-us/support/software/gastrointestinal-products/pillcam-software-v9.html}, each was extracted and processed into frames of 576 x 576 resolution. Each video was anonymized and annotated by two medical research scientist.

We randomly selected samples for 4 different lesions (whipples, ulcer, bleeding and angioectasia) in the small bowel region for the training. The entire training data consisted of 5,360 frames with an average of 1,072 for each of the four (4) categories.  We randomly split the data using 70\% for training and 30\% as the test set. During preprocessing, each frame was cropped from 576 x 576 to 500 x 500 to remove the black boundary region. We resized each frame to 224 x 224 so as to fit into the GPU memory. For each epoch, we performed augmentation using random transformations such as horizontal and vertical flip and random rotation. Our model was implemented in PyTorch \cite{NEURIPS2019_9015} and we computed the evaluation metrics using the Scikit-learn package \cite{scikit-learn}. \newline

\subsection{Experimental Results of Few-shot Lesion Recognition}
During the experiment, we performed series of test to compare performance across multiple Deep CNN architectures \cite{adewole2020deep} as well as varying the number of support samples for the few-shot recognition. While we trained the model on four (4) categories, our testing was done on five (5) different lesion categories introducing an unseen task to the model. Based on the overall aim of meta-learning, the model is expected to map the fifth category to an embedding space that is distant from the other seen categories.

We tested with three (3) different CNN architectures - VGG-19 \cite{simonyan2014very}, Resnet-50 \cite{he2016deep}, and AlexNet \cite{krizhevsky2012imagenet} - we replaced the output of the final fully connected layer with the embedding feature dimension. The parameters of the network were initialized based on pretrained ImageNet dataset \cite{russakovsky2015imagenet}.

The qualitative result of the lesion recognition performance for different CNN models is shown in table \ref{tab:performance}:

\begin{table}[h]
    \begin{center}
    \begin{tabular}{|l|c|c|c|c|}
    \hline
    \multirow{2}{*}{\textbf{Model}} & \multicolumn{4}{c|}{\textbf{Metrics}} \\
    \cline{2-5} &
     \textbf{Accuracy} &
     \textbf{Precision} & 
     \textbf{Recall} & 
     \textbf{F-score} \\ 
     \hline
    Alexnet + Triplet &
    $\textbf{{0.908}}$ &
    $\textbf{{0.914}}$ &
    $\textbf{{0.909}}$ &
    $\textbf{{0.910}}$ \\ 
    \hline
    VGG-19 + Triplet &
    ${0.893}$ &
    ${0.895}$ &
    ${0.888}$ &
    ${0.890}$ \\ 
    \hline
    Resnet + Triplet &
    ${0.854}$ &
    ${0.823}$ &
    ${0.824}$ &
    ${0.824}$ \\ 
    \hline
    \end{tabular}
    \end{center}
    \vspace{0.5cm}
    \caption{Comparison of models performance across different base CNN models}
    \label{tab:performance}
\end{table}

Table \ref{tab:performance} shows a comparative performance of the model with different base CNN models. From Table \ref{tab:performance}, we achieved best performance with the \textit{AlexNet} architecture with the \textit{VGG} trailing closely behind on all four (4) metrics.

Table \ref{tab:performance2} shows the performance of the model as we vary the number of support samples. With fewer supports, the model performance depends of the within-class variation for each category as there are fewer examples to compare with. Hard examples where there large inter-class distance will cause the model to easily make mistake. This is most common with lesions such as the angioectasia that occupies a very tiny region of the entire frame (See Figure \ref{fig:triplet_network}). With more support samples the model tries to compute a minimum over all the distances computed for each example support. 

\begin{table*}[h]
\centering

    \begin{center}
    \begin{tabular}[scale=1.7]{p{2.5cm} | p{1.7cm} | c | c | c | c | c }
    \hline
    \multirow{2}{*}{\textbf{Model}} & \multirow{2}{*}{\textbf{Metrics}} & \multicolumn{5}{c}{\textbf{Performance}} \\ 
    \cline{3-7} 
     &  & k = 1 & k = 3 & k = 5 & k = 7 & k = 9\\ 
    \hline
    \multirow{4}{*}{Alexnet + Triplet} & 
    Precision & 
    $0.826\pm 0.055$ & $0.806 \pm 0.004$ & $\textbf{0.878} \pm \textbf{0.007}$ & $\textbf{0.914} \pm \textbf{0.006}$ & $0.816 \pm 0.025$ \\ 
    & Recall & 
    $0.816 \pm 0.058$ & $0.795 \pm 0.003$ & $\textbf{0.860} \pm \textbf{0.005}$ & $\textbf{0.909} \pm \textbf{0.009}$ & $0.807 \pm 0.019$\\ 
    & F-score & 
    $0.808 \pm 0.066$ & $0.784 \pm 0.011$ & $\textbf{0.855} \pm \textbf{0.008}$ & $\textbf{0.910} \pm \textbf{0.007}$ & $0.792 \pm 0.024$ \\ 
    & Accuracy & 
    $\textbf{0.821} \pm \textbf{0.054}$ & $0.796 \pm 0.007$ & $\textbf{0.863} \pm \textbf{0.005}$ & $\textbf{0.908} \pm \textbf{0.009}$ & $0.810 \pm 0.021$ \\ 
    \hline
    \multirow{4}{*}{VGG + Triplet} & 
    Precision & 
    $0.814 \pm 0.058$ & $0.733 \pm 0.113$ & $0.808 \pm 0.015$ & $0.865 \pm 0.016$ & $\textbf{0.893} \pm \textbf{0.012}$ \\ 
    & Recall & 
    $0.796 \pm 0.069$ & $0.732 \pm 0.082$ & $0.793 \pm 0.011$ & $0.841 \pm 0.007$ & $\textbf{0.895} \pm \textbf{0.012}$\\ 
    & F-score & 
    $0.791 \pm 0.064$ & $0.715 \pm 0.096$ & $0.764 \pm 0.015$ & $0.823 \pm 0.008$ & ${\textbf{0.888} \pm \textbf{0.015}}$ \\ 
    & Accuracy & 
    $0.798 \pm 0.064$ & $0.738 \pm 0.071$ & $0.791 \pm 0.019$ & $0.841 \pm 0.005$ & $\textbf{0.890} \pm \textbf{0.012}$ \\ 
    \hline
    \multirow{4}{*}{ResNet + Triplet} & 
    Precision & 
    $\textbf{0.840} \pm \textbf{0.011}$ & $\textbf{0.825} \pm \textbf{0.013}$ & $0.822 \pm 0.027$ & $0.854 \pm 0.006$ & $0.845 \pm 0.005$ \\ 
    & Recall & 
    $\textbf{0.816} \pm \textbf{0.020}$ & $\textbf{0.803} \pm \textbf{0.012}$ & $0.801 \pm 0.030$ & $0.823 \pm 0.014$ & $0.826 \pm 0.050$ \\ 
    & F-score & 
    $\textbf{0.817} \pm \textbf{0.021}$ & $\textbf{0.797} \pm \textbf{0.011}$ & $0.802 \pm 0.033$ & \textbf{$0.824 \pm 0.012$} & $0.835 \pm 0.021$ \\ 
    & Accuracy & 
    $0.815 \pm 0.020$ & $\textbf{0.804} \pm \textbf{0.009}$ & $0.801 \pm 0.028$ & \textbf{$0.824 \pm 0.011$} & $0.817 \pm 0.053$ \\ 
    \hline
\end{tabular}
    
\end{center}
\caption{Comparison of Model performance based on k-shots}
\label{tab:performance2}
\end{table*}

From Table \ref{tab:performance2}, for a single support sample, we obtained the best performance with the \textit{ResNet} model with similar result when support was increased to three (3). With five and seven (7) supports, \textit{AlexNet} model performed better on all metrics. The \text{VGG-19} only achieved the best performance when there were more supports to compare against.
\section{Conclusion}
\label{sec:conclusion}
This work proposes a few-shot multiple lesion recognition in wireless capsule endoscopy using metric-based learning framework. Metric-based learning is designed to establish similarity or dissimilarity between concepts. Few-shot learning aims to identify new concepts from only a small number of examples. We applied these concept to multiple lesion recognition in WCE dataset.  We argued that while CNN-based classifiers have demonstrated significant improvement on object recognition and image classification task, they require vast amount of labeled dataset to train in addition to being sample inefficient. Obtaining label for WCE data is very challenging, given the volume of data and the expertise needed to provide frame-by-frame label. We approach these problems using deep metric-based learning and applied it to few-shots lesion recognition task. We experiment with different support samples as well as different base-CNN architectures. We demonstrated the effectiveness of our solution on real patients' WCE video. With our proposed solution, physicians can easily query a patient video database for specific abnormality or disease based on other clinical information. Future direction will measure the impact of different distance / similarity metric on the model performance and also extending this framework to active learning task where the model request for label on any example where it is uncertain.

\bibliographystyle{unsrt}
\bibliography{refs}  


\end{document}